\documentclass{article}

\usepackage{microtype}
\usepackage{graphicx}
\usepackage{subfigure}
\usepackage{booktabs} % for professional tables
\usepackage[pdftex,dvipsnames]{xcolor}  % Coloured text etc.
\usepackage[colorinlistoftodos,textwidth=20mm,textsize=tiny]{todonotes}

\usepackage{hyperref}
\usepackage{amsthm}

\usepackage[accepted]{icml2019}

\DeclareMathAlphabet{\mathcal}{OMS}{cmsy}{m}{n}

\newcommand{\norm}[1]{\left\lVert #1 \right\rVert}
\newcommand{\pd}[1]{p_\text{data} \left( #1 \right)}
\newcommand{\ptask}[1]{p_\text{task} \left( #1 \right)}

\newcommand{\pz}[1]{p_\vz \left( #1 \right)}

\newcommand{\bm}{\mathbf}
\newcommand{\expt}[2]{\mathbb{E}_{#1} \left[ #2 \right]}

\newcommand{\vx}{\mathbf{x}}

\newcommand{\vF}{\mathbf{F}}
\newcommand{\vG}{\mathbf{G}}

\newcommand{\vz}{\mathbf{z}}

\newcommand{\vm}{\mathbf{m}}
\newcommand{\hz}{\hat{\vz}}
\newcommand{\hx}{\hat{\vx}}

\newcommand{\lG}{\mathcal{L}_G}
\newcommand{\lF}{\mathcal{L}_F}

\newcommand{\G}{G_\theta}
\newcommand{\F}{F_\phi}
\newcommand{\D}{D_\phi}
\newcommand{\C}{C_\phi}
\newcommand{\EE}{E_\theta}

\usepackage{amsmath,amsfonts,bm}

\def\eqref#1{equation~\ref{#1}}
\def\1{\bm{1}}

\def\vm{{\bm{m}}}

\def\vx{{\bm{x}}}

\def\vz{{\bm{z}}}

\DeclareMathAlphabet{\mathsfit}{\encodingdefault}{\sfdefault}{m}{sl}
\SetMathAlphabet{\mathsfit}{bold}{\encodingdefault}{\sfdefault}{bx}{n}

\DeclareMathOperator*{\argmin}{arg\,min}

\icmltitlerunning{Deep Compressed Sensing}

\begin{document}

\twocolumn[
\icmltitle{Deep Compressed Sensing}

% It is OKAY to include author information, even for blind
% submissions: the style file will automatically remove it for you
% unless you've provided the [accepted] option to the icml2019
% package.

% List of affiliations: The first argument should be a (short)
% identifier you will use later to specify author affiliations
% Academic affiliations should list Department, University, City, Region, Country
% Industry affiliations should list Company, City, Region, Country

% You can specify symbols, otherwise they are numbered in order.
% Ideally, you should not use this facility. Affiliations will be numbered
% in order of appearance and this is the preferred way.
% \icmlsetsymbol{equal}{*}

\begin{icmlauthorlist}
\icmlauthor{Yan Wu}{to}
\icmlauthor{Mihaela Rosca}{to}
\icmlauthor{Timothy Lillicrap}{to}
\end{icmlauthorlist}

\icmlaffiliation{to}{DeepMind, London, UK}

\icmlcorrespondingauthor{Yan Wu}{yanwu@google.com}

% You may provide any keywords that you
% find helpful for describing your paper; these are used to populate
% the "keywords" metadata in the PDF but will not be shown in the document
\icmlkeywords{Machine Learning, compressed sensing, GAN, ICML}

\vskip 0.3in
]

% this must go after the closing bracket ] following \twocolumn[ ...

% This command actually creates the footnote in the first column
% listing the affiliations and the copyright notice.
% The command takes one argument, which is text to display at the start of the footnote.
% The \icmlEqualContribution command is standard text for equal contribution.
% Remove it (just {}) if you do not need this facility.

\printAffiliationsAndNotice{}  % leave blank if no need to mention equal contribution
% \printAffiliationsAndNotice{\icmlEqualContribution} % otherwise use the standard text.

\begin{abstract}
Compressed sensing (CS) provides an elegant framework for recovering sparse signals from compressed measurements.
For example, CS can exploit the structure of natural images and recover an image from only a few random measurements.
% Unlike popular autoencoding models, reconstruction in CS is posed as an optimisation problem that is separate from sensing.
CS is flexible and data efficient, but its application has been restricted by the strong assumption of sparsity and costly reconstruction process.
A recent approach that combines CS with neural network generators has removed the constraint of sparsity, but reconstruction remains slow.
Here we propose a novel framework that significantly improves both the performance and speed of signal recovery by jointly training a generator and the optimisation process for reconstruction via meta-learning.
We explore training the measurements with different objectives, and derive a family of models based on minimising measurement errors.
We show that Generative Adversarial Nets (GANs) can be viewed as a special case in this family of models.
Borrowing insights from the CS perspective, we develop a novel way of improving GANs using gradient information from the discriminator.
\end{abstract}

%\tim{Hi Yan -- a couple of quick notes.  The abstract is getting there -- I've edited a bit abstract and tried to introduce a very quick example that grounds CS for people who haven't encountered it before.  I'll look over the Intro later today.  For tables, could you make it more clear where the previous models are from and which one is `Ours'.}

\section{Introduction}

% \tim{Hi Yan: I will be on a flight for the next 6 hours.  I will be going over the abstract, intro, and discussion.  I will plan to paste my changes in when I land and get to the office.  So, perhaps leave those sections along if we want to avoid merge issues.}

% \tim{Start with a more general problem: e.g. encoding and generation are central in communication}
% \tim{In intro: give 1 full paragraph introducing CS in simple terms.  And at least 1/2 a paragraph about how CS could be put together with deep learing.}

% \mr{I think in this section you are trying to make three points: current models (such as VAEs) do end to end training, current models use the same codes for encoding and decoding, and the fact that compressed sensing does not make these assumptions, and perhaps that is a better approach. Currently these points are a bit scattered, and the intro to GANs feels a bit forced. Would creating a paragraph for each point make sense?}
% \yan{The point on end-to-end training is a bit misleading and unnecessary, so I removed it to focus on the contrast of the other two.}
Encoding and decoding are central problems in communication \cite{mackay2003information}.
Compressed sensing (CS) provides a framework that separates encoding and decoding into independent measurement and reconstruction processes \cite{candes2006stable,donoho2006compressed}.
Unlike commonly used auto-encoding models \cite{bourlard1988auto,kingma2013auto,rezende2014stochastic}, which feature end-to-end trained encoder and decoder pairs, CS reconstructs signals from low-dimensional measurements via online optimisation.
This model architecture is highly flexible and sample efficient: high dimensional signals can be reconstructed from a few random measurements with little or no training at all.
CS has been successfully applied in scenarios where measurements are noisy and expensive to take, such as in MRI \cite{lustig2007sparse}.
Its sample efficiency enables the development of, for example, the ``single pixel camera'', which reconstructs a full resolution image from a single light sensor \cite{duarte2008single}.

However, the wide application of CS, especially in processing large scale data where modern deep learning approaches thrive, is hindered by its assumption of sparse signals and the slow optimisation process for reconstruction.
Recently, \citet{bora2017compressed} combined CS with separately trained neural network generators.
% \tim{Can we say directly that they used these networks for the measurement function?}
% \yan{they actually only used NN for the generator. Since they used generators trained with both GAN and vae, I am keeping the previous sentence for now.}
Although these pre-trained neural networks were not optimized for CS, they demonstrated reconstruction performance superior to existing methods such as the Lasso \cite{tibshirani1996regression}.
Here we propose the deep compressed sensing (DCS) framework in which neural networks can be trained from-scratch for both measuring and online reconstruction.
We show that this framework leads naturally to a family of models, including GANs~\cite{goodfellow2014generative}, which can be derived by training the measurement functions with different objectives.
In summary, this work contributes the following:
% \mr{Rewrote a bit some of the below points}
\begin{itemize}
    \item We demonstrate how to train deep neural networks within the CS framework. 
    \item We show that a meta-learned reconstruction process leads to a more accurate and orders of magnitudes faster method compared with previous models.
    \item We develop a new GAN training algorithm based on latent optimisation, which improves GAN performance. The non-saturated generator loss $-\ln \left(D(G(\vz)) \right)$ emerges as a measurement error. 
    % \item We develop a new algorithm that improves the training stability of GANs. The non-saturated generator target $-\ln \left(D(G(\vz)) \right)$ is derived as a measurement error. 
    \item We extend our framework to training semi-supervised GANs, and show that latent optimisation results in semantically meaningful latent spaces. 
    % \item We develop a novel method to train conditional generative models, which results in semantically meaningful latent spaces. 
\end{itemize}

\begin{figure}
    \centering
    \includegraphics[width=0.25\textwidth]{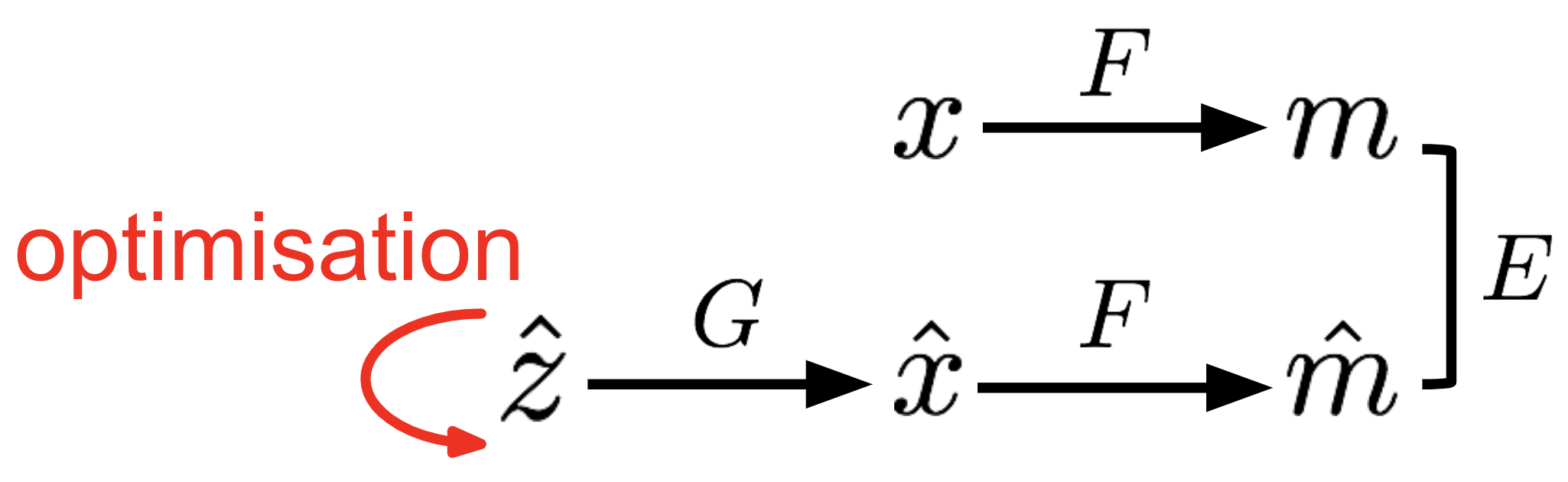}
    \caption{Illustration of Deep Compressed Sensing. $\vF$ is a measurement process that produces a measurement of the signal, $\vm$ and $\vG$ is a generator that reconstructs the signal from a latent representation $\hz$. The latent representation is optimised to minimise a measurement error $\EE(\vm, \hat{\vm})$.}
    \label{fig:model}
\end{figure}

\subsection*{Notations}

We use bold letters for vectors and matrices and normal letters for scalars. $\expt{p(\vx)}{f(\vx)}$ indicates taking the expectation of $f(\vx)$ over the distribution $p(\vx)$.
We use subscriptions of Greek letters to indicate function parameters. For example, $\G$ is a function parametrised by $\theta$.

\section{Background}

\subsection{Compressed Sensing}

Compressed sensing aims to recover signal $\vx$ from a linear measurement $\vm$:
\begin{equation}
    \vm = \vF \, \vx + \eta
    \label{eq:cs}
\end{equation}
where $\vF$ is the $C \times D$ \emph{measurement matrix}, and $\eta$ is the measurement noise which is usually assumed to be Gaussian distributed. $\vF$ is typically a ``wide'' matrix, such that $C \ll D$. As a result, the measurement $\vm$ has much lower dimensionality compared with the original signal; solving $\vx$ is generally impossible for such under-determined problems.
The elegant CS theory shows that one can nearly perfectly recover $\vx$ with high probability given a random matrix $\vF$ and sparse $\vx$ \cite{donoho2006compressed,candes2006stable}.
In practice, the requirement that $\vx$ be sparse can be replaced by sparsity in a set of basis $\Phi$, such as the Fourier basis or wavelet, so that $\Phi \, \vx$ can be non-sparse signals such as natural images.
Here we omit the basis $\Phi$ for brevity; the linear transform from $\Phi$ does not affect our following discussion.

At the centre of CS theory is the Restricted Isometry Property (RIP) \footnote{The theory can also be proved from the closely related and more general Restricted Eigenvalue condition \cite{bora2017compressed}. We focus on RIP in this form for its more straightforward connection with the training loss (see section \ref{sec:csml}).},
which is defined for $\vF$ and the difference between signals $\vx_1 - \vx_2$ as
\begin{equation}
\begin{split}
     (1 - \delta)\,\norm{\vx_1 - \vx_2}_2^2 &\leq \norm{\vF \, (\vx_1 - \vx_2) }_2^2 \\
     &\leq (1+\delta) \, \norm{\vx_1 - \vx_2}_2^2
\end{split}
\label{eq:rip}
\end{equation}
where $\delta \in (0, 1)$ is a small constant. 
The RIP states that the projection from $\vF$ preserves the distance between two signals bounded by factors of $1-\delta$ and $1 + \delta$.
This property holds with high probability for various random matrices $F$ and sparse signals $x$.
It guarantees minimising the measurement error
\begin{equation}
    \hx = \argmin_{\vx}\norm{\vm - \vF \, \vx }_2^2
    \label{eq:tcs-opt}
\end{equation}
under the constraint that $\vx$ is sparse, leads to accurate reconstruction $\hx \approx \vx$ with high probability \cite{donoho2006compressed,candes2006stable}.
% \mr{Citation}
This constrained optimisation problem is computationally intensive --- a price for the measuring process that only requires sparse random projections of the signals \cite{baraniuk2007compressive}.

\subsection{Compressed Sensing using Generative Models}
\label{sec:csgm}

The requirement of sparsity poses a strong restriction on CS.
Sparse bases, such as the Fourier basis or wavelet, only partially relieve this constraints, since they are restricted to domains known to be sparse in these bases and cannot adapt to data distributions.
Recently, \citet{bora2017compressed} proposed compressed sensing using generative models (CSGM) to relax this requirement.
This model uses a \emph{pre-trained} deep neural network $\G$ (from a VAE or GAN) as the structural constraint in the place of sparsity.
This generator maps a latent representation $\vz$ to the signal space:
\begin{equation}
    \vx = \G(\vz)
\end{equation}
Instead of requiring sparse signals, $\G$ implicitly constrains output $\vx$ in a low-dimensional manifold via its architecture and the weights adapted from data.
This constraint is sufficient to provide a generalised Set-Restricted Eigenvalue Condition (S-REC) with random matrices, under which low reconstruction error can be achieved with high probability.
A minimisation process similar to that in CS is used for reconstruction:
\begin{align}
    \hz &= \argmin_{\vz}\EE(\vm, \vz) \label{eq:argmin-csg} \\
    \EE &= \norm{\vm - \vF \, \G(\vz)}^2_2
    \label{eq:cs-err}
\end{align}
such that $\hx = \G(\vz)$ is the reconstructed signal.
In contrast to directly optimising the signal $\vx$ in CS (eq.\ref{eq:tcs-opt}), here optimisation is in the space of latent representation $\vz$.

The $\argmin$ operator in eq.~\ref{eq:argmin-csg} is intractable since $\EE$ is highly non-convex.
It is therefore approximated using gradient descent starting from a randomly sampled point $\hz \sim p_\vz(\vz)$:
\begin{equation}
\hz \leftarrow \hz - \alpha \, \frac{\partial \EE(\vm, \vz)}{\partial \vz} \bigg\rvert_{\vz=\hz}
\label{eq:z-gd}
\end{equation}
where $\alpha$ is a learning rate. One can take a specified $T$ steps of gradient descent.
Typically, hundreds or thousands of gradient descent steps and several re-starts from the initial step are needed to obtain a sufficiently good $\hat{\vz}$ \cite{bora2017compressed,bojanowski18a}.
This process is illustrated in Figure~\ref{fig:model}.

This work established the connection between compressed sensing and deep neural networks, and demonstrated performance superior to the Lasso \cite{tibshirani1996regression}, especially when the number of measurements is small.
The theoretical properties of CSGM have been more closely examined by \citet{hand2017global}, who also proved stronger convergence guarantees.
More recently, \citet{dhar2018modeling} proposed additional constraints to allow \emph{sparse deviation} from the generative model's support set, thus improving generalisation.
However, CSGM still suffers from two restrictions:
\begin{enumerate}
    \item The optimisation for reconstruction is still slow, as it requires thousands of gradient descent steps.
    \item It relies on random measurement matrices, which are known to be sub-optimal for highly structured signals such natural images. Learned measurements can perform significantly better \cite{weiss2007learning}.
\end{enumerate}
%This efficiency is insufficient for larger-scale applications.
%We address the first problem by generalise $\vM$ to a metric function and train it with gradient descent, and address the second problem using MAML inspired latent optimisation.

\subsection{Model-Agnostic Meta Learning}
\label{sec:maml}

Meta-learning, or learning to learn, allows a model adapting to new tasks by self-improving \cite{schmidhuber1987evolutionary}.
Model-Agnostic Meta learning (MAML) provides a general method to adapt parameters for a number of tasks \cite{finn2017model}.
Given a differentiable loss function $\mathcal{L}(\mathcal{T}_i; \theta)$ for task $\mathcal{T}_i$ sampled from the task distribution $\ptask{\mathcal{T}}$, the task-specific parameters are adapted by gradient descent from the initial parameters $\theta$:
\begin{equation}
    \theta_i \leftarrow \theta - \alpha \nabla_\theta \mathcal{L}(\mathcal{T}_i; \theta)
    \label{eq:inner-opt}
\end{equation}
The initial parameters $\theta$ are trained to minimise the loss across all tasks
\begin{equation}
    \min_{\theta} \expt{\mathcal{T}_i \sim \ptask{\mathcal{T}}}{\mathcal{L}(\mathcal{T}_i; \theta_i)}
    \label{eq:maml-opt}
\end{equation}
Multiple steps and more sophisticated optimisation algorithms can be used in the place of eq.~\ref{eq:inner-opt}.
Despite $\mathcal{L}$ usually being a highly non-convex function, by back-propagating through the gradient-descent process, only a few gradient steps are sufficient to adapt to new tasks.

\subsection{Generative Adversarial Networks}
\label{sec:gan-review}

A Generative Adversarial Network (GAN) trains a parametrised generator $\G$ to fool a discriminator $\D$ that tries to distinguish real data from fake data sampled from the generator \cite{goodfellow2014generative}. 
The generator $\G$ is a deterministic function that transforms samples $\vz$ from a source $\pz{\vz}$ to the same space as the data $\vx$, which has the distribution $\pd{\vx}$. %, and $\pmm{\vx}$ for the distribution of samples as outputs of the generator (model distribution).
This adversarial game can be summarised by the following min-max problem with the value function $V(\G, \D)$:
\begin{equation}
\begin{split}
    \min_{\G} \max_{\D} & V(\G, \D) = \expt{\vx \sim \pd{\vx}} {\ln \D(\vx)} \\
    & \quad+ \expt{\vz \sim p_{\vz}(\vz)}{\ln (1 - \D(\G(\vz)))}
\end{split}
\label{eq:gan_obj}
\end{equation}
GANs are usually difficult to train due to this adversarial game \cite{balduzzi2018mechanics}. Training may either diverge or converge to bad equilibrium with, for example, collapsed modes, unless extra care is taken in designing and training the model \cite{radford2015unsupervised,salimans2016improved}.

A widely adapted trick is using $-\ln \left(D(G(\vz)) \right)$ as the objective for the generator \cite{goodfellow2014generative}.
Compared with eq.~\ref{eq:gan_obj}, this alternative objective avoids saturating the discriminator in the early stage of training when the generator is too weak.
% \mr{Nit:Technically, it avoid vanishing gradients when the generator is bad and the discriminator is detecting that, that could also be in cases when it starts to learn new modes, for example}
However, this objective voids most theoretical analyses \cite{hu2018on}, since the new adversarial objective is no longer a zero-sum game (eq.~\ref{eq:gan_obj}).

In most GAN models, discriminators become useless after training. Recently, \citet{tao2018b} and \citet{azadi2018discriminator} proposed methods using the discriminator for importance sampling.
Our work provides an alternative: our model moves latent representations to areas more likely to generate realistic images as deemed by the discriminator.

\section{Deep Compressed Sensing}

We start by showing the benefit of combining meta-learning with the model in \citet{bora2017compressed}.
We then generalise measurement matrices to parametrised measurement functions, including deep neural networks.
While previous work relies on random projections as measurement functions, our approach learns measurement functions by imposing the RIP as a training objective.
% However, the same principle of minimising measurement errors for reconstruction applies.
We then derive two novel models by imposing properties other than the RIP on the measurements, including a GAN model with discriminator-guided latent optimisation, which leads to more stable training dynamics and better results.

\subsection{Compressed Sensing with Meta-Learning}
\label{sec:csml}

We hypothesise that the run-time efficiency and performance in CSGM (\citealt{bora2017compressed}, section \ref{sec:csgm}), can be improved by training the latent optimisation procedure using meta-learning, by back-propagating through the gradient descent steps \cite{finn2017model}. 
The latent optimisation procedure for CS models can take hundreds or thousands of steps.  By employing meta-learning to optimise this optimisation procedure we aim to achieve similar results with far fewer updates.
%\footnote{Although other more sophisticated optimisation method can be used for latent optimisation, we use gradient descent in our experiment as well as the description here for simplicity.}

To this end, the model parameters, as well as the latent optimisation procedure, are trained to minimise the expected measurement error:
\begin{equation}
    \min_{\theta} \, \mathcal{L}_G, \,\,  \text{for} \,
    \mathcal{L}_G = \expt{\vx_i \sim \pd{\vx}}{\EE(\vm_i, \hz_i)}
    \label{eq:cs-opt}
\end{equation}
where $\hz_i$ is obtained from gradient descent (eq.~\ref{eq:z-gd}).
The gradient descent in eq.~\ref{eq:z-gd} and the loss function in eq.~\ref{eq:cs-opt} mirror their counterparts in MAML (eq.~\ref{eq:inner-opt} and \ref{eq:maml-opt}), except that:
%\mr{First point below: I presume that you are referring to the fact the full gradient is known, while usually Monte Carlo estimation is used to estimate gradient updates. Perhaps that can be more clear?}
% \yan{I meant there's not monte carlo estimation for the gradient of z, since E depends on only one sample. I rephrased it a bit.}
\begin{enumerate}
    \item Instead of the stochastic gradient computed in the outside loop, here each measurement error $\EE$ only depends on a single sample $\vz$, so eq.~\ref{eq:z-gd} computes the exact gradient of $\EE$.
    \item The online optimisation is over latent variables rather than parameters. There are usually much fewer latent variables than parameters, so the update is quicker.
\end{enumerate}

Like in MAML, we implicitly perform second order optimisation, by back-propagating through the latent optimisation steps which compute $\hz_i$
when optimising eq. \ref{eq:cs-opt}.
We empirically observed that this dramatically improves the efficiency of latent optimisation, with only 3-5 gradient descent steps being sufficient to improve upon baseline methods.

% \mr{Maybe rephrase the sentence below? Something like Because we training the generator, etc}
% \yan{rephrased}
% Since we train the generator $\G$ (which is fixed in
Unlike~\citet{bora2017compressed}, we also train the generator $\G$.
Merely minimising eq.~\ref{eq:maml-opt} would fail --- the generator can exploit $\vF$ by mapping all $\G(\vz)$ into the null space of $\vF$.
% , violating the S-REC based on fixed generators.
This trivial solution always gives zero measurement error, but may contain no useful information.
Our solution is to enforce the RIP (eq.~\ref{eq:rip}) via training, by minimising the \emph{measurement loss}:
\begin{equation}
\begin{split}
    \mathcal{L}_F &= \expt{\vx_1, \vx_2}{\left( \norm{\vF \, (\vx_1 - \vx_2) }_2 -  \norm{\vx_1 - \vx_2}_2 \right)^2}
\end{split}
    \label{eq:rip-reg}
\end{equation}
% \mr{maybe expand on the choice of px or even simply rephrase}
% \yan{rephrased this part}
$\vx_1$ and $\vx_2$ can be sampled in various ways.
While the choice is not unique, it is important to sample from both the data distribution $\pd{\vx}$ and generated samples $\G(\vz)$, so that the trained RIP holds for both real and generated data.
In our experiments, we randomly sampled one image from the data and two generated images at the beginning and end of latent optimisation, then computed the average between the 3 pairs of losses between these 3 points as a form of ``triplet loss''. 

Our algorithm is summarised in Algorithm \ref{alg:m-cs}.
Since Algorithm \ref{alg:m-cs} still uses a random measurement matrix $\vF$, it can be used as any other CS algorithm when ground truth reconstructions are available for training the generator.

\begin{algorithm}[tb]
   \caption{Compressed Sensing with Meta Learning}
\begin{algorithmic}
   \STATE {\bfseries Input:} minibatchs of data $\{\vx_i\}_{i=1}^N$, random matrix $\vF$, generator $\G$, learning rate $\alpha$, number of latent optimisation steps $T$
   \REPEAT
   \STATE Initialize generator parameters $\theta$
   \FOR{$i=1$ {\bfseries to} $N$}
       \STATE Measure the signal $\vm_i \leftarrow \vF \, \vx_i$
       \STATE Sample $\hz_i \sim p_\vz(\vz)$
       \FOR{$t=1$ {\bfseries to} $T$}
       \STATE Optimise $\hz_i \leftarrow \hz_i - \frac{\partial}{\partial \vz} \EE(\vm_i, \hz_i)$
       \ENDFOR
%   \IF{$x_i > x_{i+1}$}
%   \STATE Swap $x_i$ and $x_{i+1}$
%   \STATE $noChange = false$
%   \ENDIF
   \ENDFOR
   \STATE $\lG = \frac{1}{N} \sum_{i=1}^N \EE(\vm_i, \hz_i)$
   \STATE Compute $\lF$ using eq.~\ref{eq:rip-reg}
   \STATE Update $\theta \leftarrow \theta - \frac{\partial}{\partial \theta} (\mathcal{L}_G + \mathcal{L}_F)$
   \UNTIL{reaches the maximum training steps}
\end{algorithmic}
\label{alg:m-cs}
\end{algorithm}

\subsection{Deep Compressed Sensing with Learned Measurement Function}

In Algorithm \ref{alg:m-cs}, we use the RIP property to train the generator. We can use the same approach and enforce the RIP property to learn the measurement function $\vF$ itself, rather than using a random projection. 

\subsubsection{Learning Measurement Function}
\label{sec:dcs_vanilla}

We start by generalising the measurement matrix $\vF$ (eq.~\ref{eq:cs}), and define a parametrised measurement function $\vm \leftarrow \F(\vx)$.
The model introduced in the previous section corresponds to a linear function $\F(\vx) = \vF \, \vx$; now both $\F$ and $\G$ can be deep neural networks.
Similar to CS, the central problem in this generalised setting is inverting the measurement function to recover the signal $\vx \leftarrow \F^{-1}(\vm)$ via minimising the measurement error similar to eq.~\ref{eq:cs-err}:
\begin{equation}
    \EE(\vm, \vz) = \norm{\vm - \F \left(\G (\vz) \right)}^2_2
    \label{eq:dcs-err}
\end{equation}
The distance preserving property as a counterpart of the RIP can be enforced by minimising a loss similar to eq.~\ref{eq:rip-reg}:
\begin{equation}
    \mathcal{L}_F = \expt{\vx_1, \vx_2}{ \left( \norm{\F(\vx_1 - \vx_2) }_2 -  \norm{\vx_1 - \vx_2}_2 \right)^2}
    \label{eq:f-rip-reg}
\end{equation}

Minimising $\mathcal{L}_F$ provides a relaxation of the constraint specified by the RIP (eq.~\ref{eq:rip}).
When $\mathcal{L}_F$ is small, the projection from $\vF$ better preserves the distance between $\vx_1$ and $\vx_2$.
This relaxation enables us to transform the RIP into a training objective for the measurements, which can then be integrated into training other model components. Empirically, we found this relaxation leads to high quality reconstruction.

The rest of the algorithm is identical to Algorithm \ref{alg:m-cs}, except that we also update the measurement function's parameters $\phi$.
Consequently, different schemes can be employed to coordinate updating $\theta$ and $\phi$, which will be discussed more in section \ref{sec:train}.
This extended algorithm is summarised in Algorithm \ref{alg:dcs}.
We call it Deep Compressed Sensing (DCS) to emphasise that both the measurement and reconstruction can be deep neural networks.
Next, we turn to generalising the measurements to properties other than the RIP.

\begin{algorithm}[tb]
   \caption{Deep Compressed Sensing}
\begin{algorithmic}
   \STATE {\bfseries Input:} minibatchs of data $\{\vx_i\}_{i=1}^N$, measurement function $\F$, generator $\G$, learning rate $\alpha$, number of latent optimisation steps $T$
   \REPEAT
   \STATE Initialize generator parameters $\theta$
   \FOR{$i=1$ {\bfseries to} $N$}
       \STATE Measure the signal $\vm_i \leftarrow \F(\vx_i)$
       \STATE Sample $\hz_i \sim p_\vz(\vz)$
       \FOR{$t=1$ {\bfseries to} $T$}
       \STATE Optimise $\hz_i \leftarrow \hz_i - \frac{\partial}{\partial \vz} \EE(\vm_i, \hz_i)$
       \ENDFOR
   \ENDFOR
   \STATE $\lG = \frac{1}{N} \sum_{i=1}^N \EE(\vm_i, \hz_i)$
   \STATE Compute $\lF$ using eq.~\ref{eq:rip-reg}
   \STATE Option 1 : joint update $\theta \leftarrow \theta - \frac{\partial}{\partial \theta} (\mathcal{L}_G + \mathcal{L}_F)$
   \STATE Option 2 : alternating update 
   \STATE \hspace{2cm} $\theta \leftarrow \theta - \frac{\partial}{\partial \theta} \mathcal{L}_G \qquad 
    \phi \leftarrow \phi - \frac{\partial}{\partial \phi} \mathcal{L}_F
   $
   \UNTIL{reaches the maximum training steps}
\end{algorithmic}
\label{alg:dcs}
\end{algorithm}
\subsubsection{Generalised CS 1: CS-GAN}
\label{sec:dcs-gan}

Here we consider an extreme case: a \emph{one-dimensional} measurement that only encodes how likely an input is a real data point or fake one sampled from the generator.
% From the coding perspective we motivated at the introduction, this example immediately requires a separate code for the generator, which needs to contain much richer information than the binary measurements to generate realistic looking samples.
% The measurements, on the other hand, only needs a very small amount of information to ensure generated samples, as generalised reconstructions, are consistent with the measured property, such as the identity of real or fake data here.
One way to formulate this is to train the measurement function $\F$ using the following loss instead of eq.~\ref{eq:f-rip-reg}:
\begin{equation}
    \mathcal{L}_F =
    \begin{cases}
    \norm{\F(\vx) - 1}_2^2 & \vx \sim p_\text{data}(\vx) \\
    \norm{\F(\hat{\vx})}_2^2 & \hat{\vx} \sim \G(\hat{\vz}), \forall \hat{\vz}
    \end{cases}
    \label{eq:lsgan_lf}
\end{equation}
Algorithm \ref{alg:dcs} then becomes the Least Squares Generative Adversarial Nets (LSGAN, \citealp{mao2017least}) with latent optimisation --- they are exactly equivalent when the latent optimisation is disabled ($T=0$, zero step).
LSGAN is an alternative to the original GAN \cite{goodfellow2014generative} that can be motivated from Pearson $\chi^2$ Divergence.
To demonstrate a closer connection with original GANs \cite{goodfellow2014generative}, we instead focus on another formulation whose measurement function is a binary classifier (the discriminator).

This is realised by using a binary classifier $\D$ as the measurement function, where we can interpret $\D(\vx)$ as the probability that $\vx$ comes from the dataset. In this case, the measurement function is equivalent to the \emph{discriminator} in GANs.
% passing the one-dimensional output of $\F$ through a sigomid function $\sigma(\vm) = \frac{1}{1 + e^{-m}}$,
% so that we can interpret $\sig{\F(\vx)}$ as the 
% To avoid cluttered notations, we use the short-hand $\D(\cdot) = \sig{\F(\cdot)}$, which also indicates \emph{discriminator} as in GANs.
Consequently, we change the the squared-loss in eq.~\ref{eq:dcs-err} to the cross-entropy loss as the matching measurement loss function \cite{bishop2006pattern} (ignoring the expectation over $\vx$ for brevity):
% \mr{A bit strange to read without expectations, but up to you}
% \yan{I will be just over the column width with expectation.. but feel free to try!}
\begin{equation}
    \mathcal{L}_F = t(\vx) \, \ln \left[ \D(\vx) \right] + (1-t(\vx)) \, \ln \left[ 1 - \D(\vx) \right]
    \label{eq:csgan_reg}
\end{equation}
where the binary scalar $t$ is an indicator function identifies whether $\vx$ is a real data point.
\begin{equation}
    t(\vx) =
    \begin{cases}
    1 & \vx \sim p_\text{data}(\vx) \\
    0 & \vx \sim \G(\vz), \forall \vz
    \end{cases}
    \label{eq:gan-indicator}
\end{equation}
Similarly, a cross-entropy measurement error is employed to quantify the discrepancy between $\D (\G (\vz))$ and the scalar measurement $m = \D(\vx)$:
\begin{equation}
\begin{split}
    \EE(m, \vz) &= m \, \ln\left[\D (\G (\vz)) \right] \\ 
    & \quad + (1-m) \, \ln \left[ 1 - \D (\G (\vz)) \right]
\end{split}
    \label{eq:csgan-err-0}
\end{equation}

At the minimum of $\mathcal{L}_F = 0$ (eq.~\ref{eq:csgan_reg}), the optimal measurement function is achieved by the perfect classifier:
\begin{equation}
    \D(\vx) =
    \begin{cases}
    1 & \vx \sim p_\text{data}(\vx) \\
    0 & \vx \sim \G(\vz), \forall \vz
    \end{cases}
\end{equation}

% Since the measurement $m = \D(\vx)$ is the probability that $\vx$ comes from the dataset, $m$ \emph{should} be $1$ given a perfect classifiers, as the measurement loss (eq.~\ref{eq:csgan_reg}) $\lF = 0$ when
% \begin{equation}
%     \D(\vx) = 1, \, \forall \vx \sim p_\text{data}(\vx)
% \end{equation}
We can therefore simplify eq.~\ref{eq:csgan-err-0} by replacing $m$ with its target value $1$ as in teacher-forcing \cite{williams1989learning}:
\begin{equation}
    E(\vm, \vz) = \ln\left[ \D (\G (\vz)) \right]
    \label{eq:csgan-err}
\end{equation}

% Comparing eq.~\ref{eq:csgan_reg} with eq.~\ref{eq:gan_obj} shows that
% now recovers the vanilla GAN with the alternative loss 
% \mr{vanilla model with altenrative loss}

% In particular, we derived the commonly used generator objective (eq.~\ref{eq:csgan-err}) as a measurement error.

This objective recovers the vanilla GAN formulation with the commonly used alternative loss~\cite{goodfellow2014generative}, which we derived as a measurement error. When latent optimisation is disabled ($T=0$), Algorithm \ref{alg:dcs} is identical to a vanilla GAN.

In our experiments (section \ref{sec:exp-gan}), we observed that the additional latent optimisation steps introduced from the CS perspective significantly improved GAN training.
We reckon this is because latent optimisation moves the representation to areas more likely to generate realistic images as deemed by the discriminator.
Since the gradient descent process remains local, the latent representations are still spread broadly in latent space, which avoids mode collapse.
Although a sufficiently powerful generator $\G$ can transform the source $\pz{\vz}$ into arbitrarily complex distribution, a more informative source, as implicitly manifested from the optimised $\vz$, may significantly reduce the complexity required for $\G$, thus striking a better trade-off in terms of the overall computation.
% \mr{Nice}

\subsubsection{Generalised CS 2: Semi-supervised GANs}
\label{sec:dcs_class}

So far, we have shown two extreme cases of Deep Compressed Sensing: in one case, the distance preserving measurements (section \ref{sec:dcs_vanilla}) essentially encode all information for recovering the original signals;
on the other hand, the CS-GAN (section \ref{sec:dcs-gan}) has one-dimensional measurements that only indicates whether signals are real or fake.
We now seek a middle ground, by using measurements that preserve class information for labelled data.

We generalise CS-GAN by replacing the binary classifier (discriminator) $\D$ with a multi-class classifier $\C$.
For data with $K$ classes, this classifier outputs $K+1$ classes with the $(K+1)$'th class reserved for ``fake'' data that comes from the generator.
This specification is the same as the classifier used in semi-supervised GANs (SGANs, \citet{salimans2016improved}).
Consequently, we extend the binary indicator function in eq.~\ref{eq:gan-indicator} to multi-class indicator, so that its $k$'the element $t^k(\vx) = 1$ when $\vx$ in class $k$.
The $k$'th output of the classifier $\C^k(\vx)$ indicates the predicted probability that $\vx$ is in the $k$'th class, and multi-class cross-entropy loss is used for the measurement loss and measurement error:
\begin{align}
    \mathcal{L}_F &= \sum_{k=1}^{K+1} t^k(\vx) \, \ln \left[ \C^k(\vx) \right] \label{eq:gc-reg}\\
    E(\vm, \vz) &= t^k(\vx) \, \ln \left[ \C^k(\G(\vz)) \right]
\end{align}
When latent optimisation is disabled ($T = 0$), the model is similar to other semi-supervised GANs \cite{salimans2016improved,acgan}.
However, when $T > 0$ the online optimisation moves latent representations towards regions representing particular classes. This provides a novel way of training conditional GANs.

Compared with conditional GANs which concatenate labels to latent variables \cite{mirza2014conditional}, optimising latent variables is more adaptive and uses information from the entire model.
Compared with Batch-Norm based methods \cite{miyato2018cgans}, the information for conditioning is presented in the target measurements, and does not need to be trained as Batch-Norm statistics \cite{ioffe2015batch}.
Since both of these methods use separate sources (label inputs or batch statistics) to provide the condition, their latent variables tend to retain no information about the condition.
Our model, on the other hand, distils the condition information into the latent representation, which results in semantically meaningful latent space (Figure~\ref{fig:samples-gc}).

\begin{table}[t]
\caption{A family of DCS models differentiated by the properties of measurements in comparison with CS. The CS measurement matrix does not need training, so it does not have a training loss.}
\label{tab:sum}
\vskip 0.15in
\begin{center}
\begin{small}
\begin{sc}
\begin{tabular}{lcc}
\toprule
Model & Property & Loss \\
\midrule
CS & RIP    & N/A \\
DCS & trained RIP    & eq.~\ref{eq:f-rip-reg} \\
CS-GAN & validity preserving & eq.~\ref{eq:csgan_reg} \\
CS-SGAN & class preserving & eq.~\ref{eq:gc-reg} \\
\bottomrule
\end{tabular}
\end{sc}
\end{small}
\end{center}
\vskip -0.1in
\end{table}

\subsection{Optimising Models}
\label{sec:train}

The three models we derived as examples in the DCS framework are summarised in Table \ref{tab:sum} along side CS.
The main difference between them lies is the training objective used for the measurement functions $\mathcal{L}_F$. 
Once $\mathcal{L}_F$ is specified, the generator objective $\lG$, in the form of measurement error, can be derived follow suit.
When $\lF$ and $\lG$ are adversarial, such as in the CS-GAN, $\F$ and $\G$ need to be optimised separately as in GANs.
This is implemented as the alternating update option in Algorithm \ref{alg:dcs}.
In optimising the latent variables (eq.~\ref{eq:z-gd}), we normalise $\hz$ after each gradient descent step, as in \cite{bojanowski18a}.
We treat the step size $\alpha$ in latent optimisation as a parameter and back-propagate through it in optimising the model loss function.
An additional technique we found useful in stabilising CS-GAN training is to penalise the distance $\vz$ moves as an optimisation cost and add it to $\mathcal{L}_G$:
\begin{equation}
    \mathcal{L}_O = \beta \cdot \norm{\hz - \vz_0}_2^2
    \label{eq:reg-transport}
\end{equation}
where $\beta$ is a scalar controlling the strength of this regulariser.
This regulariser encourages small moves of $\vz$ in optimisation, and can be interpreted as approximating an \emph{optimal transport} cost \cite{villani2008optimal}.
% When consider optimising a batch of latent $\vz$, instead just one sample, from their initial location, the problem may be better viewed as transporting one distribution of $\vz$ to another.
% A future direction is to investigate the latent optimisation in CS as an optimal transport problem.
We found a range of $\beta$ from $1.0$ to $10.0$ made little difference in training, and used $\beta=3.0$ in our experiments with CS-GAN.

% \begin{algorithm}[ht]
% \caption{Iterative signal reconstruction}
% \begin{algorithmic}
%     \REQUIRE Measurements $\vy$ of the signal, the measurement function $m$ and the generator $g$, the reconstruction step $T$
%     \STATE Randomly sample the initial embedding $\vz_0$ from $\sZ$ 
%     \FOR {t = 1 : T}  
%         \STATE The measurement error of the reconstruction $\epsilon_0 \leftarrow \norm{y - m(x_0)}^2_2$
%     \ENDFOR
%     \STATE update parameters via gradient descent to maximise $\mathcal{O}$
% \end{algorithmic}
% \label{alg:train}
% \end{algorithm}

\section{Experiments}

% \begin{table}[t]
% \caption{Summary of model parameters}
% \label{tab:params}
% \vskip 0.15in
% \begin{center}
% \begin{small}
% \begin{sc}
% \begin{tabular}{l|cc}
% \toprule
% Model & MMIST & CIFAR \\
% \midrule
% Numbe of Iterations    & 3 & 3 \\
% Optimisation Cost & 0.0 & 12 \\
% Batch size &  64  & 12 \\
% Learning rate & 0.0001 & 12 \\
% \bottomrule
% \end{tabular}
% \end{sc}
% \end{small}
% \end{center}
% \vskip -0.1in
% \end{table}

\subsection{Deep Compressed Sensing for Reconstruction}
\label{sec:exp-dcs}

We first evaluate the DCS model using the MNIST \cite{yann1998mnist} and CelebA \cite{liu2015faceattributes} datasets.
To compare with the approach in \citet{bora2017compressed}, we used the same generators as in their model.
For the measurements functions, we considered both linear projection and neural networks.
We considered both random projections and trained measurement functions, while the generator was always trained jointly with the latent optimisation process.
Unless otherwise specified, we use 3 gradient descent steps for latent optimisation. More details, including hyperparameter values, are reported in the Appendix. Our code will be available at \url{https://github.com/deepmind/deep-compressed-sensing}.

Tables \ref{tab:recons-mnist} and \ref{tab:recons-celeba} summarise the results from our models as well as the baseline model from \citet{bora2017compressed}.
The reconstruction loss for the baseline model is estimated from Figure 1 in \citet{bora2017compressed}.
DCS performs significantly better than the baseline.
In addition, while the baseline model used hundreds or thousands of gradient-descent steps with several re-starts, we only used 3 steps without any re-starting, achieving orders of magnitudes higher efficiency.
Interestingly, for fixed $F$, random linear projections outperformed neural networks as the measurement functions in both datasets across different neural network structures (row 2 and 3 of Table \ref{tab:recons-mnist} and \ref{tab:recons-celeba}).
This empirical result is consistent with the optimality of random projections described in the compressed sensing literature and the more general Johnson-Lindenstrauss lemma \cite{donoho2006compressed,candes2006stable,johnson1984extensions}.

\begin{figure}
    \centering
    \includegraphics[width=0.2\textwidth]{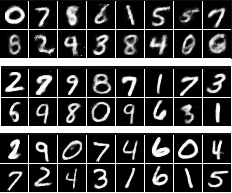}
    \caption{Reconstructions using 10 measurements from random linear projection (top), trained linear projection (middle), and trained neural network (bottom).}
    \label{fig:recons}
\end{figure}

The advantage of neural networks manifested when $\F$ was optimised; this variant reached the best performance in all scenarios.
As argued in \cite{weiss2007learning}, we observed that random projections are sub-optimal for highly structured signals such as images, as seen in the improved performance when optimising the measurement matrices (row 4 of Table \ref{tab:recons-mnist} and \ref{tab:recons-celeba}).
The reconstruction performance was further improve when the linear measurement projections were replaced by neural networks (row 5 of Table \ref{tab:recons-mnist} and \ref{tab:recons-celeba}).
Examples of reconstructed MNIST images from different models are shown in Figure \ref{fig:recons}. %\footnote{Because of potential privacy issues, we do not show the reconstructions from CelebA.}.

% UAs can be seen as an upper-bound of reconstruction performance, since they are trained directly with pixel reconstruction loss, which in our setting is considered privileged information.
Unlike autoencoder-based methods, our models were not trained with any pixel reconstruction loss, which we only use for testing. Despite this, our results are comparable with the recently proposed ``Uncertainty Autoencoders'' \cite{grover2018uncertainty}. We have worse  MNIST reconstructions: with 10 and 25 measurements, ours best model achieved 5.3 and 3.4 per-image reconstruction errors compared with theirs 3.8 and 2.5 (estimated from figure 4). However, we achieved better  CelebA results: with 20 and 50 measurements, we have errors of 23.4 and 18.5 compared with their 27 and 22 (estimated from Figure 6).

\begin{table}[t]
\caption{Reconstruction loss on MNIST test data using different measurement functions. All rows except the first are from our models. ``$\pm$'' shows the standard deviation across test samples. (L) indicates learned measurement functions. Lower is better.}
\label{tab:recons-mnist}
\vskip 0.15in
\begin{center}
\begin{small}
\begin{sc}
\begin{tabular}{l|ccc}
\toprule
Model & 10 & 25 measurements & steps \\
\midrule
% CSGM    & $54.8 \pm 3$ & $17.2 \pm 2$\\ unsure about the meaning of 'confidence interval in their paper!'
Baseline    & $54.8$ & $17.2$ & $10 \times 1000$\\
Linear& $10.8 \pm 3.8$ & $6.9 \pm 2.7$ & 3 \\
NN &  $12.5 \pm 2.2$ & $10.2 \pm 1.7$ & 3\\
% Conv & $11.8 \pm 0.5$ / $10.1 \pm 0.3$ & 12 \\
Linear(L)& $6.5 \pm 2.1$ & $4. \pm 1.4$ & 3\\
NN(L) &  $\mathbf{5.3 \pm 1.9}$ & $\mathbf{3.4 \pm 1.2}$ & 3\\
\bottomrule
\end{tabular}
\end{sc}
\end{small}
\end{center}
\vskip -0.1in
\end{table}

\begin{table}[t]
\caption{Reconstruction loss on CelebA test data using different measurement functions. All rows except the first are from our models. ``$\pm$'' shows the standard deviation across test samples. (L) indicates learned measurement functions. Lower is better.}
\label{tab:recons-celeba}
\vskip 0.15in
\begin{center}
\begin{small}
\begin{sc}
\begin{tabular}{l|ccc}
\toprule
Model & 20 & 50 measurements & steps \\
\midrule
% CSGM    & $156.8 \pm 11.8$ & $82.3 \pm 5$ \\
Baseline    & $156.8$ & $82.3$ & $2\times 500$ \\
Linear & $34.7 \pm 7.9$ & $27.1 \pm 6.1 $ & 3\\
NN &  $46.1 \pm 8.9$ & $41.2 \pm 8.3$ & 3\\
% Conv & $11.8 \pm 0.5$ / $10.1 \pm 0.3$ & 12 \\
Linear(L) & $26.2 \pm 5.9$ & $20.5 \pm 4.3$ & 3\\
NN(L) & $\mathbf{23.4 \pm 5.8}$ & $\mathbf{18.5 \pm 4.3}$ & 3\\
% Conv(L) & $5.3 \pm 0.3$ / $3.4 \pm 0.2$ & 12 \\
\bottomrule
\end{tabular}
\end{sc}
\end{small}
\end{center}
\vskip -0.1in
\end{table}

% \mr{I would just write NN (ours) for the experiments you ran using the new model}

\subsection{CS-GANs}
\label{sec:exp-gan}

To evaluate our proposed CS-GANs, we first trained a small model on MNIST to  demonstrate intuitively the advantage of latent optimisation.
For quantitative evaluation, we trained larger and more standard models on CIFAR10 \cite{krizhevsky2009learning}, and evaluate them using the Inception Score (IS) \cite{salimans2016improved} and Fréchet Inception Distance (FID) \cite{heusel2017gans}.
To our knowledge, latent optimisation has not been previously used to improving GANs, so our approach is orthogonal to existing methods such as \citet{arjovsky2017wasserstein,miyato2018spectral}.
We first compare our model with vanilla GANs, which is a special case of the CS-GAN (see section \ref{sec:dcs-gan}).

\begin{figure}
    \centering
    \includegraphics[width=0.42\textwidth]{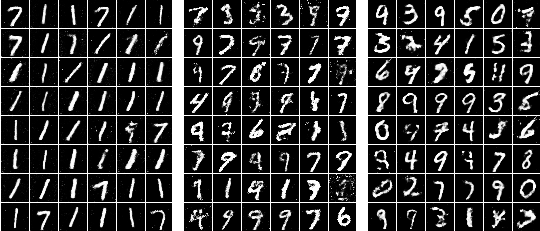}
    \caption{Samples from CS-GANs using 0 (left), 3 (central) and 5 (right) gradient descent steps in latent optimisation. The CS-GAN using 0 step was equivalent to a vanilla GAN.}
    \label{fig:samples-csgan-mnist}
\end{figure}

We use the same MNIST architectures as in section \ref{sec:exp-dcs}, but changed the the measurement function to a GAN discriminator (section \ref{sec:dcs-gan}).
We use the alternating update option in Algorithm \ref{alg:dcs} in this setting.
All other hyper-parameters are the same as in previous experiments.
We use this relatively weak model to reveal failure modes as well as advantages of the CS-GAN.
% \mr{Unclear whether same setting is even fair to look at, since they might just need different hyperparameters. Other points that can be made: for high learning rates, the GANs are unlikely to converge, if you can show that your model can reliably converge for higher learning rates, that would be very nice}
% \yan{Do you think it's fine just as an intuitive example here?}
% \mr{I think it is OK for now, but perhaphs worth changing to the learning rate analysis for the rebuttal - keeping hyperparameters constant is not fair IMO, because very different ones work for different models}
Figure \ref{fig:samples-csgan-mnist} shows samples from models with the same setting but different latent optimisation iterations. 
The three panels show samples from models using 0, 3 and 5 gradient descent steps respectively. The model using 0 iteration was equivalent to a vanilla GAN.
Optimising latent variables exhibits no mode collapse, one of the common failure modes of GAN training.
 
\begin{figure}
    \centering
     \begin{tabular}{cc}
    \includegraphics[width=0.45\columnwidth]{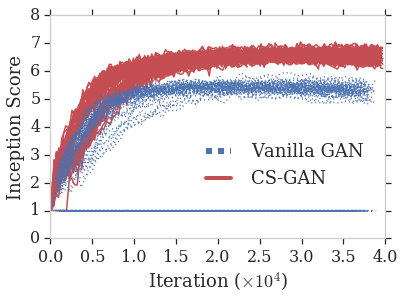} &  
    \includegraphics[width=0.45\columnwidth]{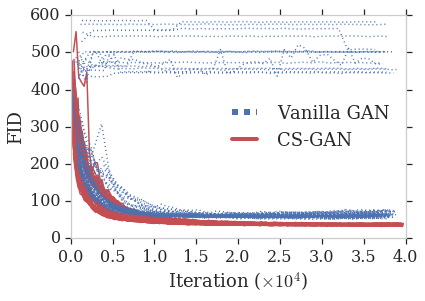}
    \end{tabular}
    \caption{Inception Score (higher is better) and FID (lower is better) during CIFAR training.}
    \label{fig:csgan-curves}
\end{figure}

To confirm this advantage, we more systematically evaluate our method across a range of 144 hyper-parameters (similar to \citet{kurach2018gan}). We use the CIFAR dataset which contains various categories of natural images, whose features from an Inception Network \cite{ioffe2015batch} are meaningful for evaluating the IS and FID.
Other than the number of gradient descent steps (0 vs.~3) the model architectures and training procedures were identical.
The change of IS and FID during training are plot in figure \ref{fig:csgan-curves}.
CS-GANs achieved better performance in both IS and FID, and had less variance across the range of hyper-parameters.
The blue horizontal lines at the bottom of Fig.~\ref{fig:csgan-curves} (left) and the top of Fig.~\ref{fig:csgan-curves} (right) shows failed vanilla GANs, but none of the CS-GANs diverged in training.

\begin{table}[t]
\caption{Comparison with Spectral Normalised GANs.}
\label{tab:sn-table}
\vskip 0.15in
\begin{center}
\begin{small}
\begin{sc}
\begin{tabular}{l|ccc}
\toprule
 & SN-GAN  & SN-GAN (ours) & CS+SN-GAN \\
\midrule
IS    & $7.42 \pm 0.08$ & $7.34 \pm 0.07$ & $\mathbf{7.80 \pm 0.05}$\\
FID & $29.3$ & $29.53 \pm 0.36$ & $\mathbf{23.13 \pm 0.50}$\\
\bottomrule
\end{tabular}
\end{sc}
\end{small}
\end{center}
\vskip -0.1in
\end{table}

We also applied our latent optimisation method on Spectral-Normalised GANs (SN-GANs) \cite{miyato2018spectral}, which use Batch Normalisation \cite{ioffe2015batch} for the generator and Spectral Normalisation for the discriminator.
%We used the same discriminator as in \citet{miyato2018spectral}, which is deeper than the DCGAN discriminator.
% We keep the same latent optimisation setting of 3 steps and initial step size $0.01$.
% The batch size was fixed to $64$, and grid search found the learning rate of $1\times 10^{-4}$ and Adam's $\beta_2$ of $0.999$ most stably achieved the best results.
We compared our model with SN-GAN in Table \ref{tab:sn-table}: the SN-GAN column reproduces the numbers from \cite{miyato2018spectral}, and the next column are numbers from our replication of the same baseline.
Our results demonstrate that deeper architectures, Batch Normalisation and Spectral Normalisation can further improve CS-GAN and that CS-GAN can improve upon a competitive baseline, SN-GAN.

\subsection{CS-SGANs}
\begin{figure}
    \centering
    \begin{tabular}{cc}
    \includegraphics[width=0.23\textwidth]{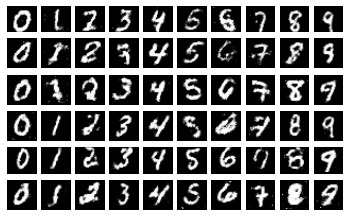} &  
    \includegraphics[width=0.21\textwidth]{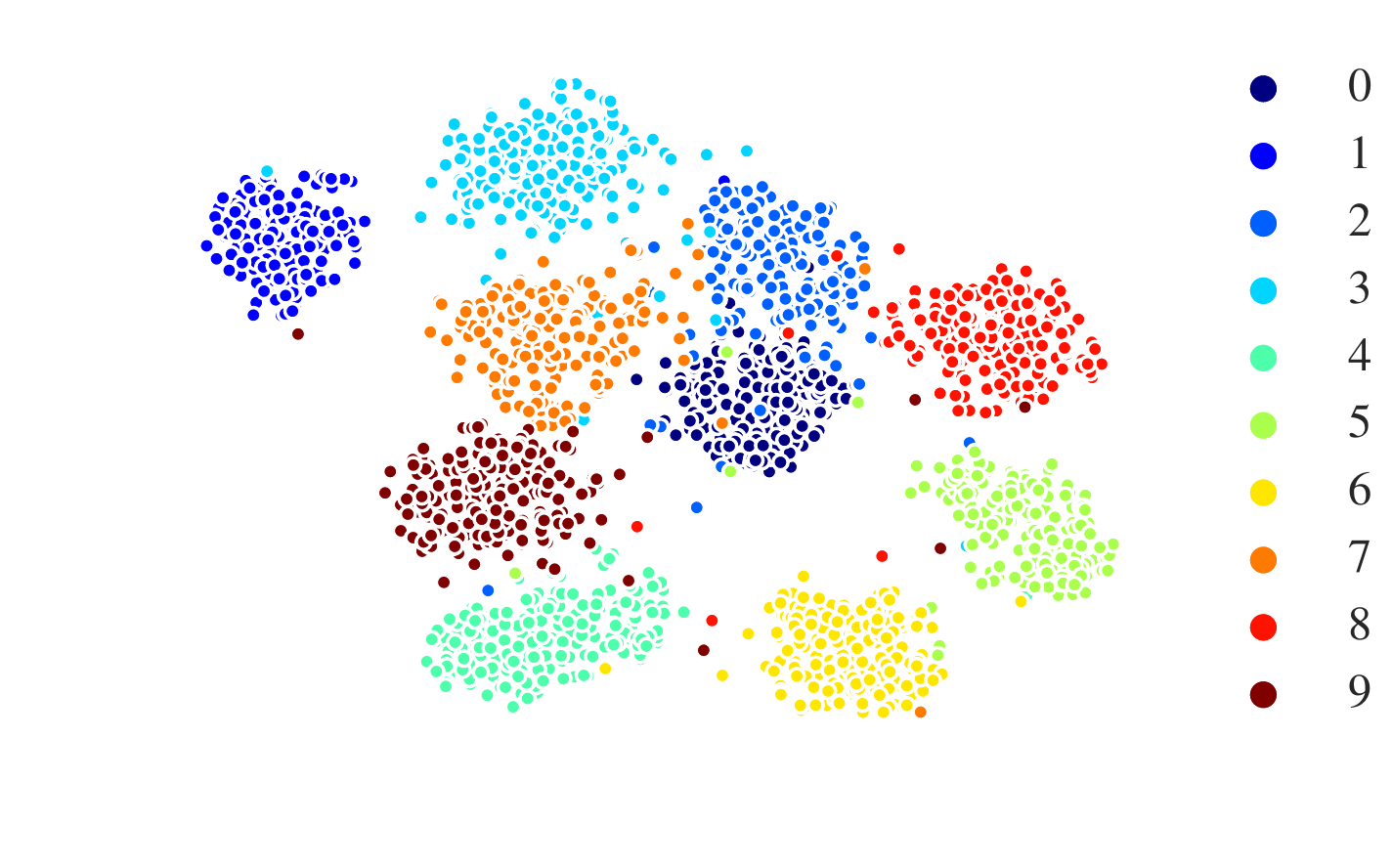}
    \end{tabular}
    \caption{Left: samples from the generative classifier, and t-SNE illustration of the generator's latent space.}
    \label{fig:samples-gc}
\end{figure}

% \mr{Consider giving a name to this as well - or improving the caption of figure 4, since it is unclear that this is something you made}
We now experimentally assess our approach to use latent optimisation in semi-supervised GANs, CS-SGAN.
We illustrate this extension with the MNIST dataset, and leave it to future work to study other applications.
% The training procedure is the same as CS-GAN, except that labels are supplied when training the measurement function.
We keep all the hyper-parameters the same as in section \ref{sec:exp-gan}, except changing the number of measurements to 11 for the 10 MNIST classes and 1 class reserved for generated samples. Samples from CS-SGAN can be seen in Figure \ref{fig:samples-gc} (left).
% shows samples from a .
% A trained generative classifier can be used as any other conditional model, for example, by directly supplying the label as a condition.
% Here we display input and output images side-by-side to highlight the more general reconstruction from CS perspective: when the class measurements only preserves class information, the ``reconstructed'' images are consistent with the class but have varying styles.
Figure~\ref{fig:samples-gc} (right) illustrates this with T-SNE~\cite{maaten2008visualizing} computed from 2000 random samples, where class labels are colour-coded.
The latent space formed separated regions representing different digits.
It is impossible to obtain such clustered latent space in typical conditional GANs \cite{mirza2014conditional,miyato2018cgans}, where labels are supplied as separate inputs while the random source only provides label-independent variations.
In contrast, in our model the labels are distilled into latent representation via optimisation, leading to a more interpretable latent space.

\section{Discussion}

We present a novel framework for combining compressed sensing and deep neural networks.
In this framework we trained both the measurement and generation functions, as well as the latent optimisation (i.e., reconstruction) procedure itself via meta-learning.
Inspired by \citet{bora2017compressed}, our approach significantly improves upon the performance and speed of reconstruction obtain in this work.
% \mr{Is the claim that this work speeds up performance and speed of reconstruction compared to Bora et all? In that case, I think the tables should make it clear what results are obtained using the method from Bora at all and point exactly to the speed up}
% \yan{Yes, bora et al. is now the 'baseline' in the tables. I made the iterations in clear in the tables now.}
In addition, we derived a family of models, including a novel GAN model, by expanding the set of properties we consider for the measurement function (Table~\ref{tab:sum}).

Our method differs from existing algorithms that aim to combine compressed sensing with deep networks in that our approach preserves the online minimisation of measurement errors in generic neural networks.
Previous attempts that combine CS and deep learning generally fall into two categories.
One category of methods interprets taking compressed measurements and reconstructing from these measurements as an encoding-decoding problem and formulate the model as an autoencoder \citep{mousavi2015deep, kulkarni2016reconnet, mousavi2017deepcodec, grover2018uncertainty, mousavi2018data,lu2018convcsnet}.
Another category of methods are designed to mimic principled iterative CS algorithms using specialised network architectures \citep{metzler2017learned,sun2016deep}.
In contrast, our framework maintains the separation of measurements from generation but still uses generic neural networks.
Therefore, both the measurements and latent representation of the generator can be flexibly optimised for different, even adversarial, objectives, while taking advantage of powerful neural network architectures.

% This flexibility of our model is manifested in CS-GAN and the generative classifier (section \ref{sec:dcs-gan} and \ref{sec:dcs_class}), where the low capacity measurements $\vm$ only preserve essential differences between subsets of data (e.g., class label, real or fake), while another much higher capacity code $\vz$ can generate high quality images with diversified styles.
% It is therefore suitable for the following common scenario: when only certain aspects of signal $\vx$ are required in communication, a sender only needs to send the highly compressed measurements $\vm$, and the receiver can decode them using their own latent code $\vz$.
% The recovered signal $\hat{\vx}$ might be different in details from $\vx$, but it would be high quality from the generative model perspective and consistent with $\vx$ in terms of the measurements $\vm$.

% Central to our work is training the measurement function, which is typically provided by random projections in classical CS.
% By expressing the desired measurement property, such as the RIP, in the form of a training objective, we can train the model, as well as the latent optimisation process, like any other parametric model.
% While it is possible to train both the measurement function and generator as deep neural networks, our work also offers the option of retaining untrained measurements (Algorithm \ref{alg:m-cs}).
% In this case, our method can be used as an efficiently reconstruction method taking the inputs into other CS algorithm as its inputs, as long as ground-truth data are available to train the generator.

Moreover, we can train measurement functions with properties that are difficult or impossible to obtain from random or hand-crafted projections, thus broadening the range of problems that can be solved by minimising error measurements online.
In other words, learning the measurement can be used as a useful stepping stone for learning complex tasks where the cost function is difficult to design directly.
% This may be done similar to what we have demonstrated here: our model achieved reconstruction from training distance preserving measurements, and achieved generation or conditional generation from training a discriminator or classifier.
Our approach can also be interpreted as training implicit generative models, where explicit minimisation of divergences is replaced by statistical tests \cite{mohamed2016learning}.
We have illustrated this idea in the context of relatively simple tasks, but anticipate that complex tasks such as style transfer \cite{zhu2017unpaired}, in areas already seen the applications of CS, as well as applications including MRI \cite{lustig2007sparse} and unsupervised anomaly detection \cite{schlegl2017unsupervised}, may further benefit from our approach.

\subsubsection*{Acknowledgments}

We thank Shakir Mohamed and Jonathan Hunt for insightful discussions. We also appreciate the feedback from the anonymous reviewers.

\appendix

\section*{Appendix}

\section{Experiments Detail}

Unless otherwise specified, we used the following default configuration for all experiments.
We used the Adam optimiser \cite{kingma2014adam} with the learning rate $1 \times 10^{-4}$ and the parameters $\beta_1 = 0.9$, $\beta_2 = 0.999$. We trained all the models for $4 \times 10^5$ steps with the batch size of $64$.
100 dimensional latent representations were used for generators.
We use 3 gradient-descent steps for latent optimisation, and the initial step size of $0.01$.

\subsection{Reconstruction Experiments}

Following \citet{bora2017compressed}, we used a 2-layer multi-layer perceptron (MLP), with 500 units in each hidden layer and leaky ReLU non-linearity, as the generator for MNIST images; for CelebA, we used the DCGAN generator \cite{radford2015unsupervised}.
In addition to random linear projections, we tested the following neural networks as the measurement functions: a 2-layer MLP with 500 units in each layer and leaky ReLU non-linearity for MNIST, and the DCGAN discriminator for CelebA.

\subsection{GAN experiments}

We used the same MLP generator and discriminator (i.e., measurement function) as described in the previous section for MNIST experiments. We also use the same architecture for the semi-supervised GAN.

For CIFAR dataset, we used the DCGAN architecture with its recommended Adam parameters $\beta_1 = 0.5$, $\beta_2 = 0.9$ \cite{radford2015unsupervised}. We tested a number of hyper-parameters as the cross product of the following: generator learning rates $\{ 1 \times 10^{-4}, 2\times 10^{-4} , 3\times 10^{-4}\}$, discriminator learning rates $\{ 1 \times 10^{-4}, 2\times 10^{-4} , 3\times 10^{-4}\}$, latent variable sizes $\{ 100 , 200\}$, mini-batch sizes $\{ 32 , 64\}$.
Additionally, 2 replicas for each combination were trained to account for the effect of random seeds.

To reproduce the Spectral Normalised GANs. We used the same discriminator as in \citet{miyato2018spectral}, which is deeper than the DCGAN discriminator.
A grid search over optimisation parameters found the learning rate of $1\times 10^{-4}$ and Adam's $\beta_2$ of $0.999$ most stably achieved the best results.

Inception Scores and Fréchet Inception Distances were reported as the averages of 10 evaluations each based on $5,000$ random samples \cite{salimans2016improved,heusel2017gans}.

\bibliography{paper}
\bibliographystyle{icml2019}

\end{document}